\documentclass[11pt,a4paper]{article}
\usepackage[hyperref]{emnlp2020}
\usepackage{times}
\usepackage{latexsym}
\usepackage{bm}
\usepackage{amsmath}
\usepackage{amssymb}
\usepackage{subcaption}
\usepackage{booktabs}

\usepackage{multirow}
\usepackage{url}

\def\Snospace~{\S{}}

\usepackage{todonotes}

\newcommand{\squishlist}{
 \begin{list}{$\bullet$}
  { \setlength{\itemsep}{0pt}
     \setlength{\parsep}{3pt}
     \setlength{\topsep}{3pt}
     \setlength{\partopsep}{0pt}
     \setlength{\leftmargin}{1.5em}
     \setlength{\labelwidth}{1em}
     \setlength{\labelsep}{0.5em} } }
     
\newcommand{\squishend}{
  \end{list}  }

\aclfinalcopy 

\title{Ensemble Distillation for Structured Prediction:\\ Calibrated, Accurate, Fast---Choose Three} 

\author{Steven Reich\textsuperscript{1} \hspace{1em} David Mueller\textsuperscript{2,3} \hspace{1em} Nicholas Andrews\textsuperscript{3}\\
  \textsuperscript{1}Department of Mathematics, University of Maryland \\
  \textsuperscript{2}Center for Language and Speech Processing, Johns Hopkins University \\
  \textsuperscript{3}Human Language Technology Center of Excellence, Johns Hopkins University \\
  \texttt{sreich47@umd.edu}
  \hspace{1em}
  \texttt{\{dam,noa\}@jhu.edu}
  }

\date{}

\begin{document}
\maketitle
\begin{abstract}
Modern neural networks do not always produce well-calibrated predictions, even when trained with a proper scoring function such as cross-entropy.
In classification settings, simple methods such as isotonic regression or temperature scaling may be used in conjunction with a held-out dataset to calibrate model outputs.
However, extending these methods to structured prediction is not always straightforward or effective; furthermore, a held-out calibration set may not always be available.
In this paper, we study \emph{ensemble distillation} as a general framework for producing well-calibrated structured prediction models while avoiding the prohibitive inference-time cost of ensembles.
We validate this framework on two tasks: named-entity recognition and machine translation. We find that, across both tasks, ensemble distillation produces models which retain much of, and occasionally improve upon, the performance and calibration benefits of ensembles, while only requiring a single model during test-time.
\end{abstract}

\section{Introduction}

\begin{figure*}[th!]
    \centering
    \subfloat[]{{\includegraphics[width=0.9\columnwidth]{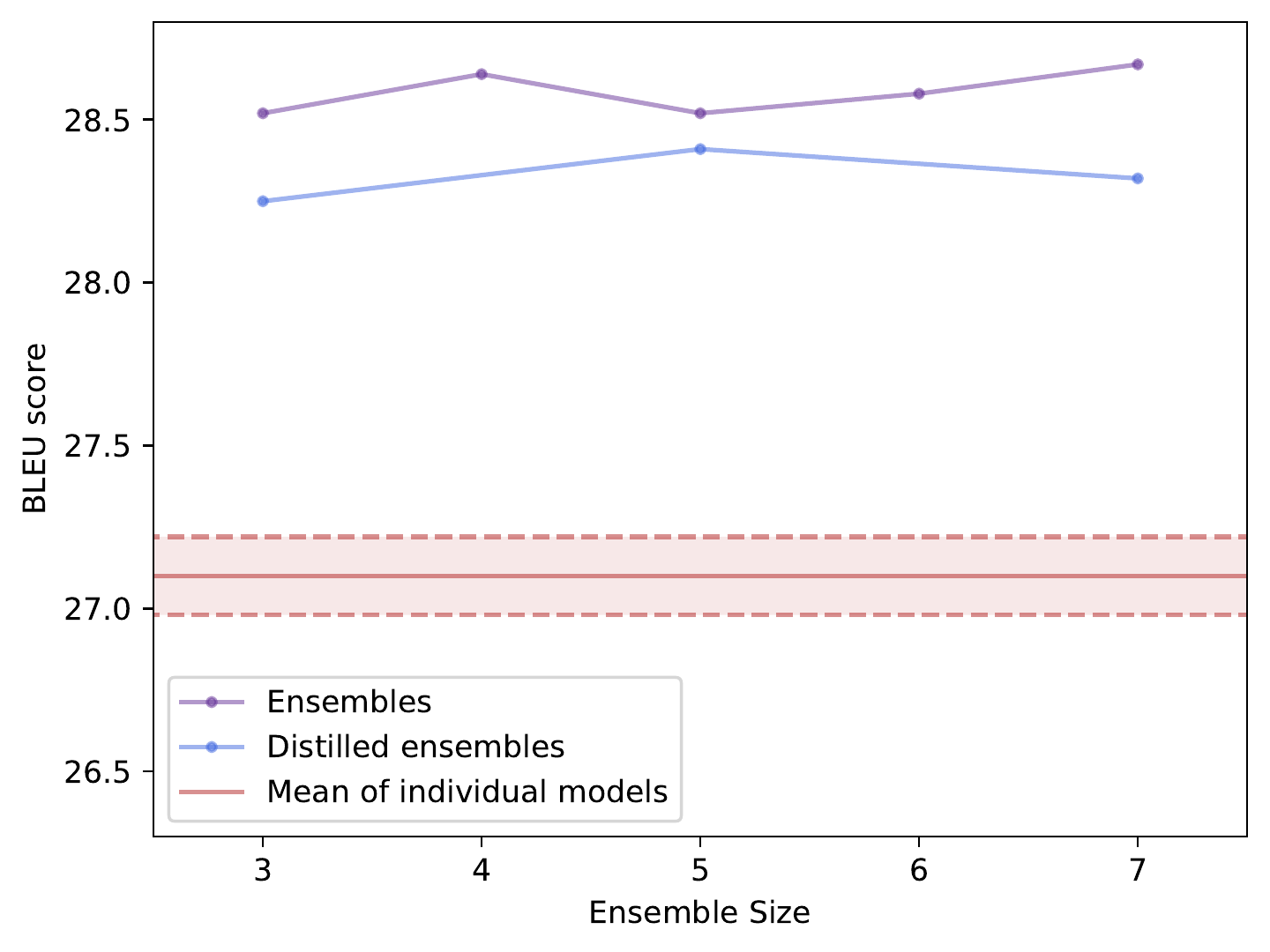} }}
    \hspace{0.5cm}
    \subfloat[]{{\includegraphics[width=0.9\columnwidth]{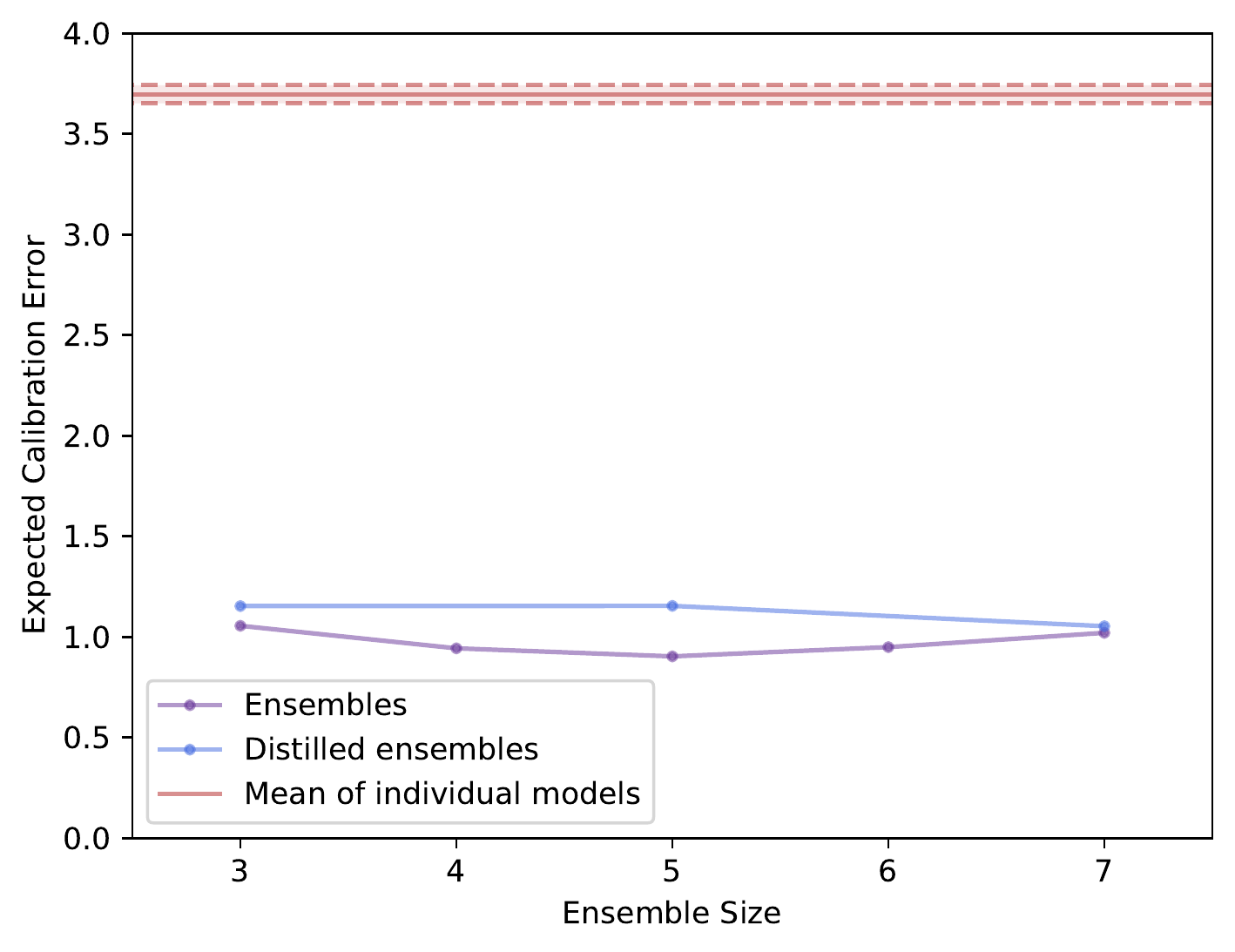} }}
    \caption{Plots of a) BLEU score ($\uparrow$ is better) and b) Top-1 ECE ($\downarrow$ is better) of ensembles and distilled ensembles of NMT models, compared to the mean (and standard error, the shaded region) of five individual standard models. Ensembles vastly improve both performance and calibration over individual models, and ensemble distillation is able to retain much of this improvement in a single model. Further, we find that even small ensembles, e.g.\ of size 3, are enough to see significant improvements over single models. Experimental details are described in~\autoref{sec:nmt}.}
    \label{fig:resultfig}
\end{figure*}

For a calibrated model, an event with a forecast confidence $p$ occurs in held-out data with probability $p$. Calibrated probabilities enable meaningful decision making, either by machines such as downstream probabilistic systems~\citep{nguyen-oconnor-2015-posterior}, or by end-users who must interpret and trust system outputs~\citep{Jiang2012CalibratingPM}.
The calibration of modern neural models has recently received increased attention in both the natural language processing and machine learning communities.
A major finding is that modern neural networks do not always produce well-calibrated predictions~\citep{guo2017calibration}. 
As a result, much recent work has focused on improving model calibration, predominantly with \emph{post-hoc} calibration methods~\citep{kuleshov2015calibrated}.


However, post-hoc calibration methods have primarily been developed in the context of classification tasks. Thus, it is unclear how these methods will affect the performance of sequence-level structured prediction tasks~\cite{kumar2019calibration}.
Additionally, post-hoc calibration methods require a held out calibration dataset, which may not be available in all circumstances.
To improve calibration, an alternate approach is model ensembling, which is closely related to approximating the intractable posterior distribution over model parameters~\cite{lakshminarayanan2017simple,pearce2018uncertainty,49462}.
Although computationally expensive, both at training and inference time, ensembling does not require a separate calibration set.
Furthermore, ensembles have been found to be competitive or even outperform other calibration methods, particularly in more challenging settings such as dataset shift~\cite{snoek2019can}.

In this paper, we study \emph{ensemble distillation} as a means of achieving calibrated and accurate structured models while avoiding the high cost of naive ensembles at inference time~\citep{hinton2015distilling}. Ensemble distillation consists of two stages: In the first stage, we select a base model for the task, such as a recurrent neural network or Transformer, and then train an ensemble of $K$ such models, ensuring diversity either via sub-sampling (\autoref{sec:ner}) or with different random seeds (\autoref{sec:nmt}). In the second stage, the ensemble of $K$ teacher models is distilled into a single student model.
Prior work has examined the effects of ensemble distillation on measures of uncertainty in vision tasks~\citep{li2019reducing,englesson2019efficient}.
To our knowledge, this is the first systematic study of the effect of ensemble distillation on the calibration of structured prediction models---we consider NER and NMT---which we find poses distinct
challenges both in terms of measuring calibration and efficiently distilling large ensembles.

To this end, our contributions may be summarized as follows:
\squishlist
\item
Our key finding is that a model distilled from an ensemble consistently outperforms baseline single models (\autoref{sec:ner}, \autoref{sec:nmt}), both in terms of calibration \emph{and} task performance.

\item
We propose a straightforward memoization technique which, when combined with a top-K approximation, enables distillation of large ensembles with negligible training overhead for NMT (\autoref{sec:nmt-ensembles}).
\item
We study the interaction between ensembling, distillation, and other commonly employed techniques including stochastic weight averaging and label smoothing in NMT (\autoref{sec:nmt-discussion}).
\item
We investigate methods to produce effective ensembles in structured prediction settings, finding that small numbers of independent models initialized from different random seeds outperform an alternative based on single optimization trajectories (\autoref{sec:single-model-ens}).
\item
Finally, we compare the calibration performance of ensembles relative to temperature scaling, which requires a separate calibration dataset, finding that it provides an orthogonal benefit (\autoref{sec:temp-scaling}).
\squishend

Our findings suggest that ensemble distillation has potential to become a standard training recipe in settings where calibration is important.


\section{Calibration}\label{sec:cal}


Given an arbitrary observation $X$ and a model with parameters $\theta$, we are interested in the \textit{predictive uncertainty}, $p_\theta(Y \mid X)$, of an event $Y$.
Our objective is to compute the predictive uncertainty of $p_\theta$ over of a finite sample of held-out data, $\{(X^{(i)}, Y^{(i)})\}_{i=1}^N$ of size $N$.
We then say that $p_{\theta}$ is \textbf{calibrated} if the predictive uncertainty agrees with held-out observations; that is, if the model predicts an event with confidence $p$, then that event prediction is correct $p\%$ of the time.

Calibrated models can be useful for down-stream systems which benefit from accurate estimates of uncertainty \citep{Jiang2012CalibratingPM, nguyen-oconnor-2015-posterior}. Recently, it has been noted that a large portion of modern neural networks are not well calibrated after training \citep{nguyen-oconnor-2015-posterior, Ott2018AnalyzingUI, kumar2019calibration}, although it has been found that pre-training can help with this in natural language processing \citep{desai2020calibration}.

\subsection{Measuring calibration}\label{sec:measure-calib}

In this work, we are interested in tasks where $Y=\{y_1, y_2, \ldots, y_T\}$ is a sequence, such that each $y_i$ is drawn from some fixed vocabulary $\mathcal{V}$ such as a fixed set of named-entity types or a language-specific sub-word vocabulary.\footnote{Note that this encapsulates a wealth of ``sequence-to-sequence'' problems of interest such as sequence tagging, translation, co-reference resolution, and (linearized) parsing.} 
However, due to the combinatorially large size of the output space $\mathcal{Y}$, any event $Y\in\mathcal{Y}$ has a minuscule probability, making it difficult to meaningfully calculate calibration. 
Thus, when evaluating the calibration of $p_\theta$, we focus on calibration with respect to token-level sub-sequences of $Y$, i.e. $p_\theta(y_t \mid X)$.

Since we evaluate model calibration on a finite amount of data, it is not possible to directly determine exactly what proportion of all events with probability $p_\theta$ will be correct.
Instead, various metrics have been proposed to estimate how well calibrated a model is.
Our evaluations in this work center around two metrics which are common in the literature:
the \emph{Brier score} \citep{brier-score}, which is the mean squared error between the model's predictions and the targets,
and the \emph{Expected Calibration Error} \citep[ECE;][]{bayesian-binning}, which uses binning to measure the correlation between confidence and accuracy.
Following \cite{nguyen-oconnor-2015-posterior}, we use \emph{adaptive binning} to select bin boundaries that allow an equal number of sampled confidences per bin.

\subsection{Addressing calibration}

A number of \emph{post-hoc} solutions to the problem of poor calibration have been proposed, including Platt scaling \citep{Platt99probabilisticoutputs}, isotonic regression \citep{iso-reg}, and temperature scaling \cite{guo2017calibration}.
However, these methods were predominantly designed for classification problems; in structured prediction problems, post-hoc re-calibration can sometimes hurt original performance \cite{kumar2019calibration}.
Additionally, post-hoc methods assume the availability of a held-out calibration set, which may not always be feasible in some settings.
Thus, improving neural network calibration during the training procedure is still an open area of research.

It is well-known that neural model ensembles may improve task performance
relative to individual models, although at the cost of increased compute and memory resources during training and inference~\cite{simonyan2014deep, 10.1109/ICCV.2015.123, jozefowicz2016exploring}.
Recently, it has been observed that ensembles of independent models trained with different random seeds also manifest improved calibration~\cite{lakshminarayanan2017simple,snoek2019can}.
Intuitively, independently initialized models may be over- or under-confident in different ways on ambiguous inputs; as a result, the average of their predictive distributions provides a more robust estimate of the true uncertainty associated with any given input.


\section{Ensemble Distillation}

\subsection{Knowledge distillation}

\citet{hinton2015distilling} first proposed knowledge distillation as a procedure to train a low-capacity \emph{student} model on the fixed distribution $q$ of a higher-capacity \emph{teacher} model.
In its general form, the distillation loss $\mathcal{L}_{\text{Student}}$ optimized by the student model with parameters $\theta$ has the form
\begin{align*}
\mathcal{L}_{\text{Student}}(\theta) = (1 - \beta) \; * \; &\mathcal{L}_{\text{NLL}}(\theta, \mathcal{D})  \\ 
+ \; \beta \; *\;&\mathcal{L}_{\text{KD}} (p_\theta, q, \mathcal{D}),
\end{align*}
where $\beta$ is an interpolation between the standard negative log-likelihood loss\footnote{Possibly against target distributions which have been augmented by label smoothing.} ($\mathcal{L}_\text{NLL}$) and the knowledge distillation loss ($\mathcal{L}_{\text{KD}}$), and $\mathcal{D}$ is the training dataset.
In general, $\mathcal{L}_{\text{KD}}$ is some measure of dissimilarity
between a the student and teacher distributions over examples in the training data, typically cross-entropy or KL-Divergence. 


As our full output space $\mathcal{Y}$ is combinatorially large, exact comparison of $p_\theta(\mathcal{Y} \mid X)$ and $q(\mathcal{Y} \mid X)$ is intractable.
A common method to address this is to instead distill teacher distributions at the \emph{token level} \citep{hinton2015distilling,kim-rush-2016-sequence}.
In models that make Markov-assumptions, such as some NER models with CRF layers, we can efficiently compute the token-level distributions marginalized over all possible label sequences $\mathcal{Y}$ for each token.
In auto-regressive models, such as the NMT models we consider, marginalization over all possible sequences is intractable. 
In this case, the token-level loss is evaluated using teacher-forcing \citep{teacher-forcing}, by conditioning on true targets up to time $t$.

\subsection{Ensemble distillation}\label{sec:framework}

Ensemble distillation uses knowledge distillation to train a student model on the output of an ensemble.
Most previous approaches to ensemble distillation collapse the ensemble distribution into a single point estimate by averaging the teachers' outputs~\citep{hinton2015distilling,bayesian-dark-knowledge,kuncoro2016distilling}.
This has been shown to be an effective way of distilling the uncertainty captured by an ensemble in computer vision tasks~\citep{li2019reducing,englesson2019efficient}.
Recently \citet{Malinin2020Ensemble} showed that by instead distilling the \emph{distribution} over the ensemble into a \emph{prior network}~\citep{prior-networks}, the student can learn to model both the \emph{epistemic} and \emph{aleatory} uncertainty of the ensemble.

As our goal is to improve model calibration, which captures both types of uncertainty, we follow previous methods of ensemble distillation which collapse the ensemble distribution into a point-estimate by uniformly averaging the distributions of each teacher. 
Formally, given an ensemble of $K$ models, our task is to train a single student model to match a teacher distribution $q$ which is composed of the $K$ distributions from the ensemble, $q_k$.
Maintaining consistency with how we derive predictions from an ensemble, when performing token-level distillation we construct the teacher distribution $q$ as a mixture of each ensemble distribution:
\begin{align*}
    q( y_t \mid X) &= \frac{1}{K}\sum_{k=1}^K q_k( y_t \mid X )    
\end{align*}
In addition to token-level distillation, \citet{kim-rush-2016-sequence} proposed \textit{sequence-level} distillation, which approximates the global distribution with the top $M$ samples and treats each samples as an additional training example during student learning.
This technique can be prohibitively expensive to use, as it increases the training time of the student by a factor of $M$;
a problem which is exacerbated during ensemble distillation, as the factor becomes $M\times K$.
To maintain simplicity in our distillation procedure, and comparability to tasks for which this technique does not apply,\footnote{For example, an NER model with a Conditional Random Field, for which we can already obtain globally normalized token-level posterior distributions.} we focus only on token-level ensemble distillation.

\section{Ensemble Distillation for NER}\label{sec:ner}

\begin{table*}[t]

\centering
\scalebox{0.80}{
\begin{tabular}{c c | c c c c c || c c c c c }\toprule
\multicolumn{2}{c}{}&\multicolumn{10}{c}{\textbf{IID Model}} \\
\multicolumn{2}{c}{}& \multicolumn{5}{c}{\textit{English}}&\multicolumn{5}{c}{\textit{German}}\\
\midrule
Model & Setting & BS+ & BS- & B-BS & B-ECE & F1  & BS+ & BS- & B-BS & B-ECE & F1 \\ 
\midrule
\multirow{2}{*}{Individual} 
&Avg & 6.401 & 0.310 & 6.109 & 5.516 & 91.11 & 17.212 & 0.233 & 9.444 & 8.542 & 81.68 \\
&$\pm$ & 0.068 & 0.006 & 0.102 & 0.048 & 0.05 & 0.359 & 0.006 & 0.144 & 0.064 & 0.26 \\
\hline
\multirow{3}{*}{Ensemble}&3 & 5.693 & 0.249 & 4.946 & 3.031 & 91.64 & 15.544 & 0.169 & 8.306 & 5.343 & 82.73 \\
&6 & \textbf{5.539} & \textbf{0.241} & 4.862 & \textbf{2.863} & \textbf{91.76} & 14.615 & 0.167 & 8.058 & 5.016 & \textbf{83.53} \\
&9 & 5.451 & \textbf{0.241} & \textbf{4.852} & 3.017 & 91.74 & \textbf{14.457} & \textbf{0.158} & \textbf{8.015} & \textbf{4.337} & 83.51 \\ \midrule
\multirow{2}{*}{Ensemble}&3 & 5.801 & 0.269 & 5.246 & 3.744 & 91.49 & 13.920 & 0.230 & 8.824 & 6.739 & 82.60\\
\multirow{2}{*}{Distillation}&6 & 5.936 & 0.256 & 5.144 & 3.683 & 91.61 & 14.440 & 0.228 & 8.958 & 6.763 & 82.02 \\
&9 & 5.959 & 0.259 & 5.174 & 3.289 & 91.51 & 14.495 & 0.197 & 8.699 & 5.694 & 82.86 \\
\midrule
\multicolumn{2}{c}{}&\multicolumn{10}{c}{\textbf{CRF Model}} \\
\multicolumn{2}{c}{}& \multicolumn{5}{c}{\textit{English}}&\multicolumn{5}{c}{\textit{German}}\\
\midrule
Model & Setting & BS+ & BS- & B-BS & B-ECE & F1  & BS+ & BS- & B-BS & B-ECE & F1 \\ 
\midrule
\multirow{2}{*}{Individual} 
&Avg & 6.998 & 0.308 & 5.911 & 4.607 & 90.37 & 15.942 & 0.233 & 8.997 & 6.047 & 80.60 \\
&$\pm$ & 0.153 & 0.012 & 0.094 & 0.092 & 0.11  & 0.256 & 0.005 & 0.089 & 0.095 & 0.14 \\
\hline
\multirow{3}{*}{Ensemble} &3 & 6.078 & 0.271 & 5.334 & 3.539 & 91.30 & 15.653 & 0.194 & 8.432 & 4.994 & 81.56 \\
&6 & 5.939 & 0.243 & 4.932 & \textbf{2.179} & 91.46 & 15.446 & 0.185 & 8.485 & 4.485 & 81.37 \\
&9 & 5.872 & \textbf{0.235} & \textbf{4.811} & 2.219 & \textbf{91.52} & 15.629 & 0.176 & 8.375 & 4.313 & 81.47 \\
\midrule
\multirow{2}{*}{Ensemble}&3 & 6.055 & 0.261 & 5.113 & 3.574 & \textbf{91.52} & 15.155 & \textbf{0.161} & \textbf{7.877} & \textbf{3.936} & 82.87 \\
\multirow{2}{*}{Distillation}&6 & \textbf{5.545} & 0.268 & 5.086 & 3.451 & 91.13 & \textbf{14.582} & 0.171 & 8.178 & 3.956 & \textbf{83.13} \\
&9 & 5.874 & 0.286 & 5.464 & 4.259 & 91.46 & 15.273 & 0.164 & 8.240 & 4.056 & 81.95 \\
\bottomrule
\end{tabular}
}
\caption{\label{table:ner-results} Ensemble and Ensemble-Distillation results on CoNLL NER. All values are percentages. \textbf{Bold} results represent the best results of each model (IID or CRF) for each metric. Note that ensembles have higher F1 and are better calibrated than individual models. Furthermore, the distilled ensemble also significantly outperforms single models in all metrics. Surprisingly, distilling token-level CRF distributions can boost student models \emph{past} the ensembles abilities. Dev results for these experiments are in \autoref{sec:ner-dev}.}
\end{table*}

We evaluate the calibration and performance effects of ensemble distillation on NER models.
In these experiments, we examine teacher ensembles that use either strong independence assumptions (subsequently referred to as ``IID''), or 1st order Markov assumptions.
These settings allow us to examine the effects of distilling \emph{globally}-marginalized versus \emph{locally}-marginalized structured distributions into student models. 
We experiment on the 2003 CoNLL Dataset \citep{tjong-kim-sang-de-meulder-2003-introduction}, which contains datasets in English and German, and consider both languages in our experiments.

Our NER models use representations from pre-trained masked language models: BERT for English and multilingual-BERT (mBERT) for German \cite{devlin-etal-2019-bert}.
Given an input sequence $X=\{x_1, \ldots, x_T\}$  BERT outputs representations for each $x_t$.\footnote{We fine-tune only the last 4 layers of BERT and mBERT. All other BERT parameters are frozen.}
We consider two separate models in our experiments: The `IID' model makes predictions based solely off of the token-level logits output from a feed-forward layer applied to the BERT representations, making each prediction $\hat y_t$ independently from all others.
The `CRF' model instead passes the representations through a bi-directional LSTM layer, and the result into a conditional random field (CRF) with learned transition parameters~\citep{lample-etal-2016-neural}.  All models are trained using the Adafactor optimizer; we use a learning rate of $1e\text{--}4$ for training the ensembles, and $5e\text{--}5$ during distillation.

For each dataset, we trained $K=9$ models in both the IID and CRF framework. To encourage diversity, each model in a given framework uses a different 1/10 split\footnote{This leaves a split of 1/10th of training data which acts as a unique validation set for the distillation training.}
of the training set for its early stopping criterion, in addition to using a different random seed.
We then consider ensembles of sizes $K = 3, 6, 9$ models, where for $K = 3, 6$ the individual models are chosen randomly, but in such a way that the ensemble of $3$ is always a subset of the ensemble of $6$.
During inference time, each ensemble's per-token distribution $q_k(y_t \mid X)$ is averaged uniformly to create the ensemble's distribution $q(y_t \mid X)$.
In IID ensembles, $q_k(y_t \mid X)$ is taken directly from the logits at timestep $t$. In the CRF model, the Forward-Backward algorithm is used to compute distributions for each timestep which are globally normalized over all possible output sequences $\mathcal{Y}$.
Predictions are then made for each token based on the maximum likelihood prediction from the ensembled distribution, $q$.
For comparison, we also train a collection of 9 models, each with a different random initialization, on the \emph{entire} training dataset and report their average performance and standard error. Note that this setup \emph{disadvantages} the ensemble, as each individual model in an ensemble has access to strictly less training data than the individual models.

\subsection{Ensemble distillation}

During ensemble distillation, we only distill \emph{into} IID models, although we consider both IID and CRF ensembles as teacher distributions.
This allows us to examine the effects of distilling globally marginalized distributions into locally marginalized models.
Each student's distillation loss $\mathcal{L}_\text{KD}$ is the token-level cross-entropy\footnote{We did not use label smoothing, which is not commonly used in NER, for the results here, as we found to not improve results. Details can be found in \autoref{sec:label-smoothing}.} between the 
student's distribution $p_\theta(y_t \mid X)$ and the ensembled distribution $q(y_t \mid X)$, with an interpolation parameter of $\beta = \frac{5}{6}$ between $\mathcal{L}_\text{KD}$ and the true train loss, $\mathcal{L}_\text{NLL}(\theta)$.
All distilled models are trained using the final 1/10th training split as validation data.

\subsection{Evaluating calibration in NER}\label{sec:nercal}

For highly imbalanced data, like NER labels, common measurements of calibration do not sufficiently distinguish between models \cite{benedetti2010scoring}. One way we account for this is to use \textit{stratified Brier score} \citep{wallace2014improving}, which has two components: the Brier score over all positive events (BS$^+$) and over all negative events (BS$^-$), whereby one of these (usually BS$^+$) is more sensitive to a model's calibration.

However, we note a potential drawback of relying on BS$^+$, namely that it is entangled with the model's recall.\footnote{That is, lower recall is strongly correlated to worse BS$^+$.} We also wish to use Expected Calibration Error, which more closely captures calibration in the sense defined in \autoref{sec:cal}, but ECE is also rendered useless in an imbalanced setting.

To address both of these issues, we therefore propose an alternative ``balanced'' version of each metric: for each entity-type,\footnote{We collapse B and I tags into type-level annotations for this measurement.} we consider the top $2N$ most confident predictions, where $N$ is the number of tokens with true labels of that class.
After this filtering, ``Balanced ECE'' (B-ECE) is computed as the weighted sum of each class's (adaptively binned) ECE.
``Balanced Brier score'' (B-BS) is similarly computed as the weighted sum of each class's Brier score over this filtered set.
These metrics correct the problem of imbalance and better reflect a model's calibration independent of its recall (and thus test performance).

\subsection{Results}

We report the results of single models, ensembles, and distilled ensembles on F1, BS-, BS+, B-BS, and B-ECE in Table~\ref{table:ner-results}. We find that, across all settings and languages, ensembles outperform individual models in both F1 and calibration. Distillation only moderately hurts these numbers. Distilled models still outperform single models; additionally, they are vastly better calibrated than single models, indicating that distillation is effective at retaining the calibration benefits of ensembles.

While distilling IID ensembles into an IID student generally lowers performance compared to the ensemble, distilling the ensembled CRF distributions obtained into an IID model can actually yield higher performance \emph{and} calibration than the ensemble. 
This suggests that global CRF distributions may not ensemble well at the token level, but are still effective distillation teachers when distilled into a local IID model with no global distribution of its own.

As a further examination of the benefits of improved calibration, we produce precision-recall curves (PR) by thresholding token-level probabilities. We find that improved calibration translates to higher area under the PR curve. Figures and experimental details are reported in~\autoref{sec:thresholds}.

\section{Ensemble Distillation for NMT}\label{sec:nmt}

\begin{table*}[tb]
\centering

\begin{subtable}[h]{.48\textwidth}
\centering
\caption{English $\rightarrow$ German}
\scalebox{0.73}{
\begin{tabular}{@{}llcccc@{}}\toprule
{\bf Method} & {\bf Model}   & {\bf \# Models} & {\bf BLEU} & {\bf ECE-1} & {\bf ECE-5} \\ \midrule
\multirow{5}{*}{LS + SWA} & \multirow{2}{*}{Individual} & Avg               & 27.45        & 3.466       & 1.295      \\
                          && $\pm$           & 0.05         & 0.078       & 0.010      \\  \cmidrule{2-6}

      &\multirow{3}{*}{Ensemble}  & 3               & 28.40        & 5.979       & 1.673      \\
                          && 5               & 28.74        & 6.611       & 1.758      \\
                          && 7               & 28.82        & 6.810       & 1.796      \\ \midrule
\multirow{8}{*}{CE + SWA\;\;} & \multirow{2}{*}{Individual} & Avg               & 27.10        & 3.697       & 1.149      \\ 
                          && $\pm$           & 0.12         & 0.045       & 0.015      \\ \cmidrule{2-6}
        &\multirow{3}{*}{Ensemble}                 & 3               & 28.52        & 1.055       & 0.313      \\
                          && 5               & 28.52        & 0.920       & 0.301      \\
                          && 7               & 28.67        & 1.020       & 0.328      \\ \cmidrule{2-6}
                          &\multirow{2}{*}{Distilled}  & 3               & 28.25	       & 1.153	     & 0.527      \\
&\multirow{2}{*}{\;\;$V'$: 64}                          & 5               & {\bf 28.41}        & 1.154       & {\bf 0.507}      \\
                          && 7               & 28.32        & {\bf 1.053}       & 0.588      \\
\bottomrule
\end{tabular}}
\end{subtable}
\hfill
\begin{subtable}[h]{.48\textwidth}
\centering
\caption{German $\rightarrow$ English}
\scalebox{0.73}{
\begin{tabular}{@{}llcccc@{}}\toprule
{\bf Method} & {\bf Model}   & {\bf \# Models} & {\bf BLEU} & {\bf ECE-1} & {\bf ECE-5} \\ \midrule
\multirow{5}{*}{LS + SWA} & \multirow{2}{*}{Individual} & Avg  & 30.98 & 2.330 & 1.198      \\
                          && $\pm$  & 0.029 & 0.016 & 0.003              \\  \cmidrule{2-6}

      &\multirow{3}{*}{Ensemble}  & 3 & 32.46  & 4.891  & 1.513                \\
                                 && 5 & 32.95  & 5.442  & 1.617                   \\
                                 && 7 & 32.98  & 5.714  & 1.663                   \\ \midrule
\multirow{8}{*}{CE + SWA\;\;} & \multirow{2}{*}{Individual} & Avg & 30.57 & 4.92 & 1.50           \\ 
                          && $\pm$  & 0.033 & 0.013 & 0.006             \\ \cmidrule{2-6}
        &\multirow{3}{*}{Ensemble}   & 3  & 32.23 & 1.941 & 0.603                    \\
                                    && 5  & 32.45 & 1.612 & 0.459                   \\
                                    && 7  & 32.74 & 1.496 & 0.401                   \\ \cmidrule{2-6}
&\multirow{2}{*}{Distilled}  & 3 & 31.71 & 1.519 & {\bf 0.591}                   \\
&\multirow{2}{*}{\;\;$V'$: 64} & 5 & 31.63 & {\bf 1.456} & 0.659                            \\
                          && 7 & {\bf 31.84} & 1.497 & 0.636                \\
\bottomrule
\end{tabular}}
\end{subtable}

\caption{\label{table:nmt-results} Performance of  Transformer-Base ensembles and individual models  on the WMT16 English $\rightarrow$ German (a) and German $\rightarrow$ English (b) tasks. ECE values are given as percentages. LS+SWA and CE+SWA indicate models trained with and without label smoothing, respectively.
Additionally, we report the performance of distilling CE-SWA ensembles into a single student model (see~\autoref{sec:nmtdetails} for details).
Similar to our NER results, we find that distillation is able to retain much of the benefits of an ensemble, both in terms of performance and calibration, over individual models.
The best \emph{single-model} performance is in {\bf bold}.
}
\end{table*}

In this section, we evaluate ensemble distillation for NMT models.\footnote{All NMT experiments are run using the \texttt{fairseq} framework~\cite{ott-etal-2019-fairseq}, using standard recipes and commodity hardware.} 
State-of-the-art NMT
models such as the Transformer are auto-regressive, meaning that the probability
of a given target $y_t$ is a function of all previous targets $y_{<t}$~\cite{vaswani2017attention}.
Thus, distilling teacher information in this scenario is different from what is done in NER; the structure level knowledge which is being distilled is inherently greedy (the teacher distributions do not take into account future sequences) and the distributions are built off of the gold labeled sequences up to that point (making it difficult to distill the global behavior of the ensemble).

All experiments are run on the WMT16 En$\rightarrow$De and De$\rightarrow$En tasks, using the vanilla Transformer-Base architecture from \citet{vaswani2017attention}.\footnote{Unless otherwise specified, our experimental configuration mirrors that of \citet{vaswani2017attention} model with the ``base'' architecture.}
We use a vocabulary of 32K symbols based on a
joint source and target byte pair encoding~\cite{sennrich2015neural,ott2018scaling}.
Unlike in our NER setting, \emph{all} models are trained on the full training set, with variation being instilled only through different random initializations and data order.
All models considered use stochastic weight averaging \citep[SWA;][]{weight-averaging}.
Additionally, to evaluate the effect of label smoothing~ \citep{label-smoothing, muller2019does} on calibration we consider 2 variations of NMT experiments: Models trained on standard cross-entropy loss (CE-SWA), and models trained on cross-entropy loss with a smoothing factor of $\lambda = 0.1$ (LS-SWA).
Models are added to ensembles based on order of random seed.\footnote{For example, an ensemble of 4 models will contain the models trained with random seeds 1, 2, 3, and 4. Note that this is essentially random selection, but it enforces that all ensembles are strict subsets of larger ensembles.} During ensemble inference, the next output token is taken from the argmax of the averaged token-distribution across all models in the ensemble.

\subsection{Challenges of ensemble distillation}\label{sec:nmt-ensembles}

Token-level distillation requires access to the teacher distribution during training, which in our experiments involves a distribution over 32K subwords. As we are interested in distilling an \emph{ensemble} of teacher models, when training on devices like GPUs with limited memory, it may not feasible to keep all models in the ensemble on device. Even on devices with sufficient memory, the additional overhead associated with ensemble inference may lead to impractical training times.

To enable scaling to large ensembles with minimal training overhead, we memoize to disk the ensemble predictive distributions associated with \emph{each token in the training data}. During training, the memoized values are streamed along with source and target subwords for calculation of the NLL and distillation losses.
However, this solution incurs a large storage cost, namely $O(T \cdot V)$ floating point numbers for a training dataset consisting of $T$ tokens and $V$ subwords.
Thus, we propose the following simple approximation scheme to reduce the storage requirements to $O(T \cdot V')$, where $V' \ll V$. 
For each token $t = 1, \ldots, T$, we store a vector $\bm{v}^{(t)} \in \mathbb{Z}^{V'}$ of indices associated with the top-$V'$ tokens of the teacher distribution, along with a vector $\bm{p}^{(t)} \in \mathbb{R}^K$ of corresponding probabilities. During training, $\mathcal{L}_{\text{KD}}$ is evaluated with respect to these fixed top-$V'$ events.

\subsection{Distillation experimental details}\label{sec:nmtdetails}

As we found label smoothing to significantly hurt ensemble calibration (\autoref{table:nmt-results}), our distillation experiments only consider the CE-SWA ensembles as teachers.
We use a truncation level of $V' = 64$ in~\autoref{table:nmt-results} and report additional results for different truncation amounts in~\autoref{table:nmt-truncation}.  
The distillation loss with weight $\beta$ is evaluated over the tokens which are in top-$V'$ using a fixed temperature of $1$.
The negative log-likelihood loss with weight $1 - \beta$ is identical to other models and also uses label smoothing with $\lambda = 0.1$.
All results use a weight of $\beta = 0.5$ on the distillation objective and use a random initialization of the model parameters, which preliminary experiments suggested was optimal.\footnote{In preliminary experiments, we also explored other training strategies, such as initializing from a constituent model using a larger weight of $\beta=0.9$ on the distillation objective. However, this did not work as well as training from scratch
with $\mathcal{L}_{\text{NLL}}$ and $\mathcal{L}_{\text{KD}}$ evenly weighted.}
Other experimental details match those of single models.

\subsection{Results}\label{sec:nmt-discussion}

Calibration for NMT is typically measured using the ECE of next-token predictions (ECE-1).\footnote{See \citep{muller2019does,kumar2019calibration}.} To better understand the calibration of the \emph{distribution} of the model's predictions, we supplement this with the ECE of the top five predictions at each token (ECE-5).\footnote{We use adaptive binning, as described in section~\autoref{sec:measure-calib}, to compute both metrics.} We report the BLEU scores and calibration metrics of our ensembles, students, and baseline models in Table~\ref{table:nmt-results}.

\begin{figure}[t]
    \centering
    \subfloat[Models trained with label smoothing]{\includegraphics[width=\columnwidth]{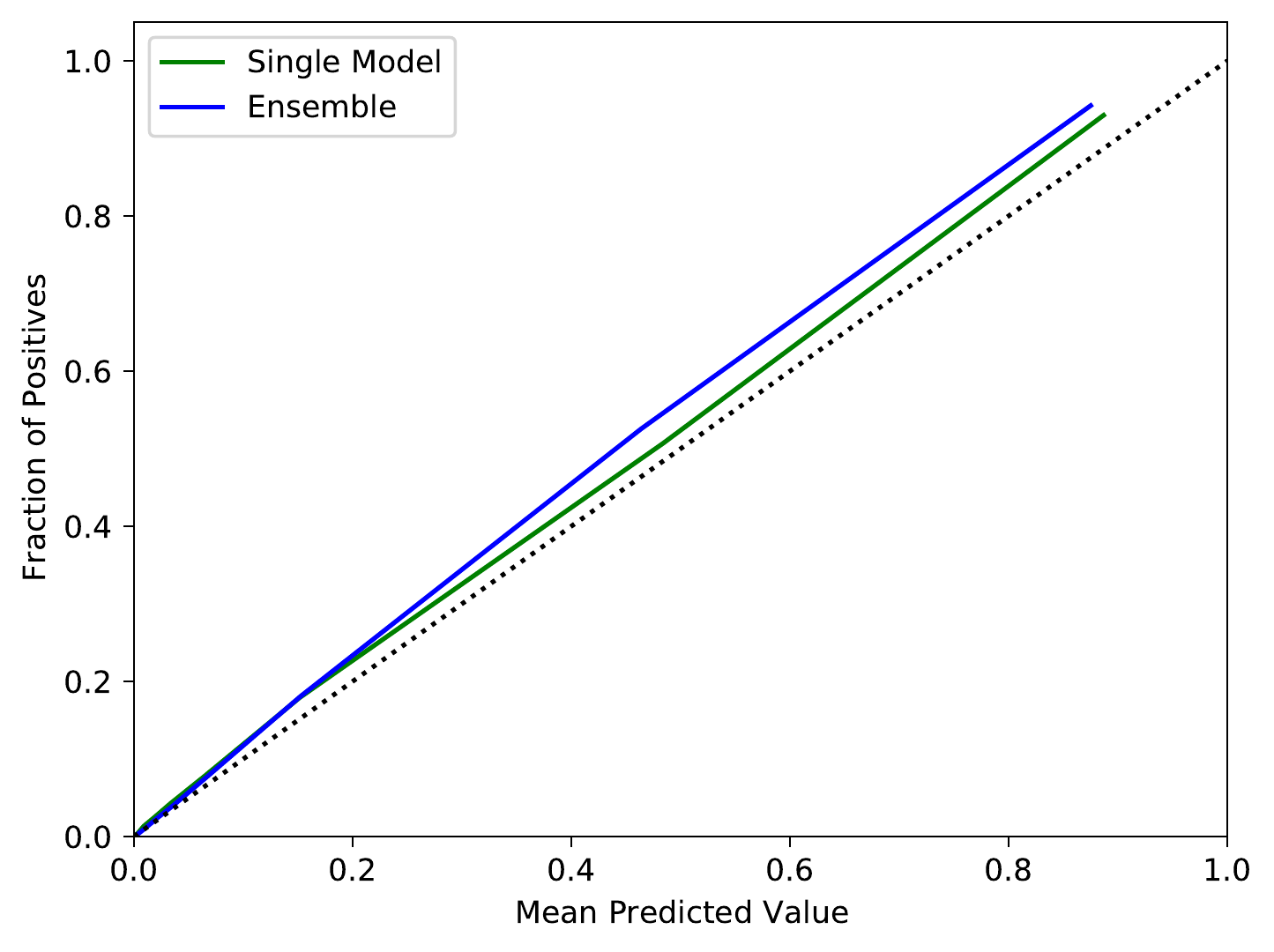} }
    \qquad
    \subfloat[Models trained without label smoothing]{\includegraphics[width=\columnwidth]{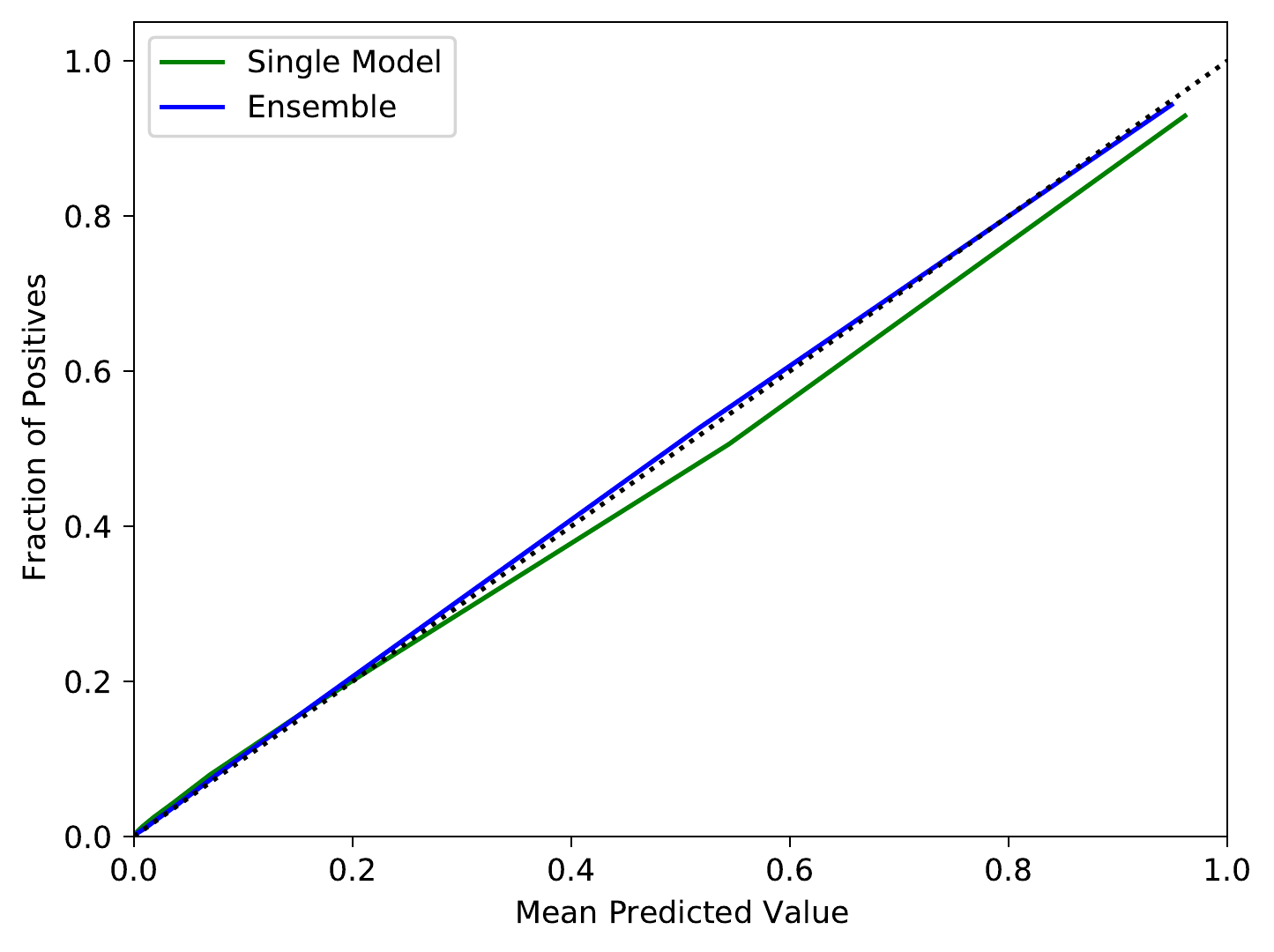} }
    \caption{Reliability plots comparing predictions of single models to those of ensembles for models trained a) with and b) without label smoothing. The dotted diagonal line represents perfect calibration; the regions above and below it correspond to underconfidence and overconfidence, respectively.}
    \label{fig:rel-plots}
\end{figure}

We find that individual models trained with label smoothing have slightly better BLEU scores and calibration than those trained without, which is consistent with the findings in \cite{muller2019does}, in which they attribute this improvement to reducing overconfidence. Surprisingly, however, ensembles of models trained using label smoothing actually have \emph{worse} calibration than independent models, and this effect grows as more models are incorporated. We hypothesize that penalizing overconfidence is effective for improving calibration of a single model, but that this results in overcorrection when models which have been similarly penalized are ensembled together. This is supported by the reliability plots in \autoref{fig:rel-plots},  which show that the individual LS models are under-confident in their top predictions, which is compounded by ensembling, whereas non-LS individual models are slightly overconfident in their top predictions, which is corrected by ensembling.

For ensembles that do not incorporate label smoothing, we observe the same trends for NMT as we do for NER: ensembles consistently improve performance, and distillation results in a single model which significantly outperforms baseline models both in terms of calibration and BLEU. We also see a more consistent trend of improvement as the ensemble size increases, which we attribute to the substantially larger NMT dataset size (\autoref{fig:resultfig}). 

\begin{table}[t]
\centering
\scalebox{0.87}{
\begin{tabular}{@{}cccc@{}}\toprule
     ${\bf V'}$          & {\bf BLEU} & {\bf ECE-1} & {\bf ECE-5} \\ \midrule
     32                & 31.64      & 1.602       & {\bf 0.623}         \\
     64                & {\bf 31.84}      & 1.497       & 0.636      \\
     128               & 31.80      & 1.567       & 0.642         \\ 
     256               & 31.72      & {\bf 1.325}       & 0.648 \\
     \bottomrule
\end{tabular}}
\caption{\label{table:nmt-truncation} Distillation performance for De$\rightarrow$En as the truncation $V'$ is varied. An ensemble of 7 models is used as the teacher.}
\end{table}

\noindent \textbf{Effect of truncation size ${\bf V'}$}. We consider $V' = \{32, 64, 128, 256\}$ which requires $\{32, 64, 128, 254\}$ gigabytes of storage respectively to memoize the teacher distributions. To put these storage requirements in perspective, naively storing the full predictive distribution would require approximately 17 terabytes of storage. Note that the storage requirements of the proposed distillation procedure are constant with respect to the number of models in the teacher ensemble, so in principle the proposed approach could be used to distill significantly larger ensembles than considered in this work. The results for De$\rightarrow$En  are reported in~\autoref{table:nmt-truncation}. Surprisingly, as $V'$ becomes smaller, performance does not monotonically degrade, suggesting that truncation could have a beneficial regularisation effect.
In fact, although $V'=32$ has a marginally worse BLEU score than the best models, it has the best ECE-5 score. This suggests that for large datasets it may be reasonable to use aggressive truncations, although we do not experiment with values smaller than $V'=32$.

\section{Further Experiments}

\subsection{Single-model ensembles}\label{sec:single-model-ens}

Our findings suggest that even ensembles of relatively small size (3-4) can still yield significant improvements over single models.
In this section, we explore whether these findings can be mirrored by an ensemble which is built from a \emph{single optimization trajectory}, built from multiple checkpoints.

For this purpose we consider a popular technique introduced by \citet{loshchilov2016sgdr}. The authors define SGDR, a scheme for training with a cyclical learning rate, and find that an ensemble of `snapshots' of the model taken when the learning rate is at a minimum gives similar improvements in accuracy to proper ensembles.

We follow the same procedure used to train our single CE+SWA NMT models, stopping 3 epochs earlier. We then warm-start this model and train for 3 epochs\footnote{We use 3 epochs to align with the recommendation in \cite{loshchilov2016sgdr}. For SGDR, we set $T_{\text{mult}}$ parameter to 1, but found that other settings gave similar results. Reported runs used 2000 steps between saved models, but values in $\{500, 1000, 3000\}$ did not produce significantly different results.} using a cyclical learning rate, saving the model at the end of each. 
\autoref{table:nmt-singlerun} gives the results obtained by ensembling the saved checkpoints, and a comparison to an equivalent proper ensemble. We find that SGDR improves calibration over single models, but not to the same extent as the ensemble, and does not improve BLEU. 

\begin{table}[t]
\centering
\scalebox{0.87}{
\begin{tabular}{@{}lcccc@{}}\toprule
     {\bf Method}          & {\bf \# Models} & {\bf BLEU} & {\bf ECE-1} & {\bf ECE-5} \\ \midrule
     SGDR                  & 3               & 26.97      & 1.105       & 0.394         \\
     CE + SWA              & Ind.             & 27.10        & 3.697       & 1.149      \\
     CE + SWA              & 3               & \bf{28.52} & \bf{0.904}  & \bf{0.303}         \\ 
     \bottomrule
\end{tabular}}
\caption{\label{table:nmt-singlerun} Single-run ensemble performance for NMT. We include the performance of the 3-model ensemble and the average individual model performance. We find that single-run ensembles have better calibration than single models, but do not see the same performance gains that true ensembles do.}
\end{table}

Applying SGDR to NER experiments yielded results which did not improve over the individual NER models in \autoref{table:ner-results}. We posit that using pretrained BERT reduces the amount of diversity which can be introduced in a single training run. 

\subsection{Temperature scaling}\label{sec:temp-scaling}

One of the benefits of the proposed framework is that it does not require the use of a separate validation set to achieve improvements in calibration. This also means that when one is available, it can be used in conjunction with our method to further improve calibration. A well-studied method for performing post-hoc re-calibration using additional data is temperature scaling \cite{guo2017calibration}. 
To explore the interactions between temperature scaling and ensemble distillation, we perform temperature scaling on our German NER IID models and our largest IID ensemble, using the validation set for tuning calibration. Additionally, we train a new student on the temperature-scaled ensemble.
We report the test performance and calibration of all models, compared to the models without temperature scaling, in~\autoref{table:ner-tempscale}.

We find that temperature scaling can improve individual model calibration, but it does not surpass the calibration of ensembles.\footnote{Note that temperature scaling has no effect on an individual IID model's performance, as it does not change the ranking of predictions.}
Additionally, we see that temperature scaling can further be used to improve the calibration of both ensembles and ensemble-distilled models. However, the effect on performance varies; while temperature scaling hurts ensemble performance, it has a significant positive effect on the student model.


\begin{table}[t]
\centering
\scalebox{0.7}{
\begin{tabular}{l c | c c c c c }\toprule
Model & TS & BS+ & BS- & B-BS & B-ECE & F1 \\
\midrule
\multirow{2}{*}{Individual} && 17.212 & 0.233 & 9.444 & 8.542 & 81.68 \\
 &\checkmark & 16.463 & 0.184 & 8.052 & 4.098 & 81.68 \\
\hline

\multirow{2}{*}{Ensemble: 9} &&14.457 & 0.158 & 8.015 & 4.337 & 83.51 \\
&\checkmark& 14.806      & 0.141 & 7.978      & 3.428      & 83.32 \\
\hline

\multirow{2}{*}{Distilled: 9} && 14.495 & 0.197 & 8.699 & 5.694 & 82.86 \\
&\checkmark & 14.081      & 0.160      & 7.857      & 3.647      & 84.27 \\
\bottomrule
\end{tabular}}
\caption{\label{table:ner-tempscale} CoNLL-2003 German IID results for individual models, 9-ensembles, and distilled 9-ensembles with and without temperature scaling (TS). We find that we can utilize temperature scaling in all cases to boost calibration, but temperature scaling only helps overall performance when used in combination with distillation.}
\end{table}

\section{Conclusion}\label{sec:conclusion}

{\bf Summary of contributions}. We present a systematic study of the effect of ensembles on the calibration of structured prediction models, which consistently improve calibration and performance relative to single models.
Our key finding is that \emph{ensemble distillation} may be used to produce a single model that preserves much of the 
improved calibration \emph{and} performance of the ensemble while being as efficient as single models at inference time.
Furthermore, we show that calibration of the single student models can be further improved by other, orthogonal, re-calibration methods.
We release all code and scripts.\footnote{\url{https://github.com/stevenreich47/ensemble-distillation-structured-prediction}}

\vspace{5pt}

\noindent {\bf Open research questions}. Non-autoregressive translation (NAT) is an active area of research for NMT~\cite{gu2017non,stern2019insertion,ghazvininejad2019mask}. Most knowledge distillation for NAT is performed at the \emph{sequence level}, and ignores distributional information at the token level.
In future work, we are interested in exploring NAT using distilled ensembles with truncated distributions, and assessing how improved calibration impacts non-sequential decoding performance.
Finally, \citet{snoek2019can} find that deep ensembles can significantly improve out-of-domain performance over single models, and we are interested in exploring whether our distillation techniques retain these benefits. 

\section*{Acknowledgements}

This work was supported, in part, by the Human Language Technology Center of Excellence at Johns Hopkins University.
We would like to thank the anonymous reviewers for their insightful comments and critique.
We would also like to thank Suzanna Sia and Kevin Duh for their feedback and questions about this work.

\bibliography{emnlp2020}
\bibliographystyle{acl_natbib}

\clearpage
\appendix
\onecolumn

\section{The effect of calibration on PR curves for NER}\label{sec:thresholds}

In this section, we illustrate a further advantage of calibrated NER models, which is that they enable straightforward thresholding of the returned confidences at different operating points of interest. In general, one may be willing to trade-off precision or recall according to the application. The popular F1 metric for NER evaluates at one such operating point. The framework of precision-recall (PR) curves provides a graphical illustration of performance of different models across a range of operating points, and the \emph{area under the PR curve} provides a summary statistic that enables comparing different models across the entire range of operating points~\cite{flach2015precision}.

Note however that sequence distributions do not enable straightforward thresholding because the probability of any particular sequence is vanishingly small. Therefore, it is necessary to consider marginal probabilities of positions or short spans instead. While expensive in the case of the CRF, requiring dynamic programming for each possible event of interest, note that our distilled IID model enables direct thresholding on the \emph{calibrated} per-position probabilities.

\begin{figure}[h]
\centering
\scalebox{1.0}{
    \includegraphics{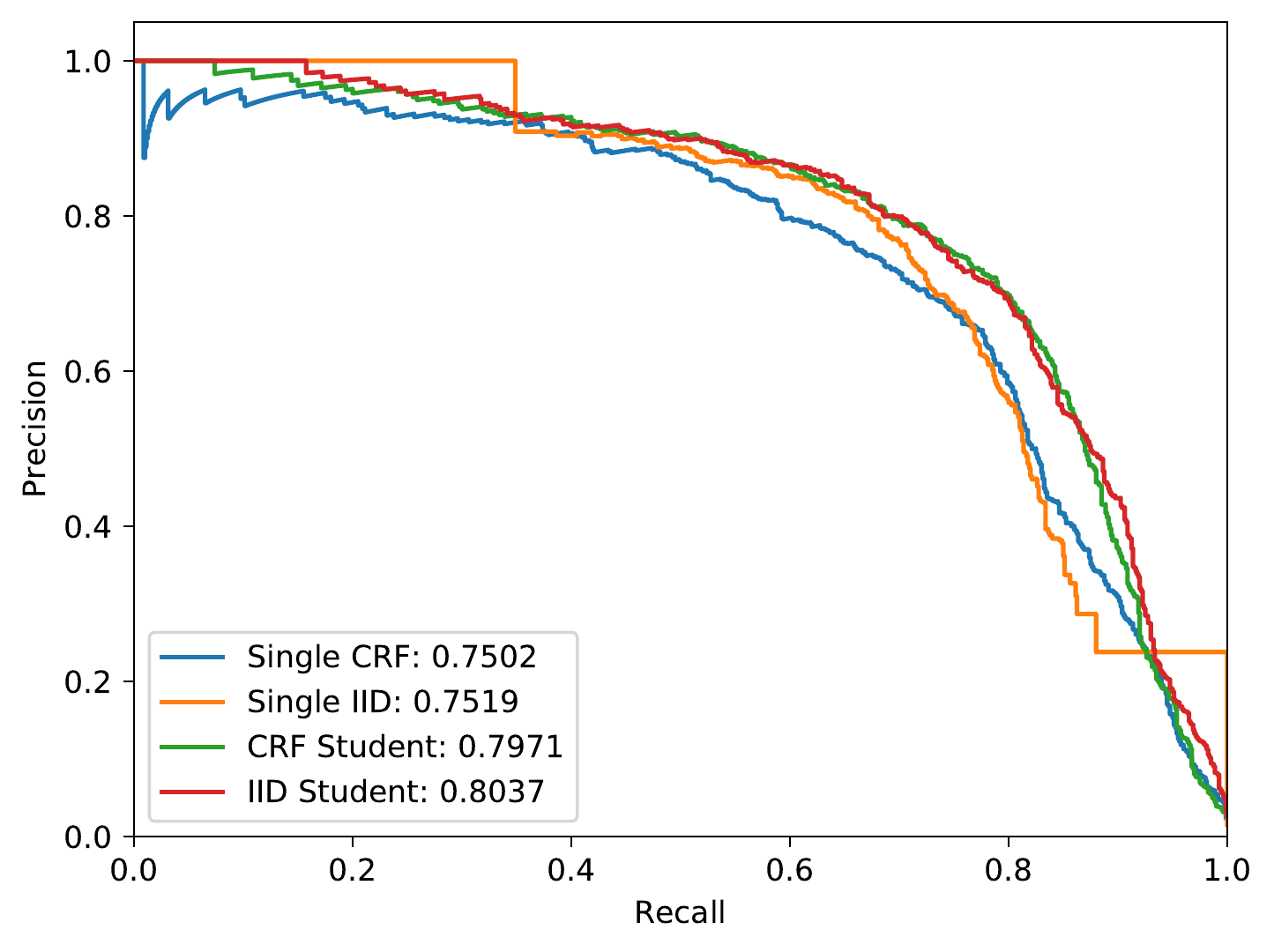}
    }
    \caption{Precision-Recall curves of NER models.}
    \label{fig:pr-curves}
\end{figure}

To illustrate the benefit of improved calibration, \autoref{fig:pr-curves} shows PR curves for four models:
\begin{itemize}
\item An individual CRF model
\item An individual IID model
\item A model distilled from a 9-ensemble of CRFs
\item A model distilled from a 9-ensemble of IIDs
\end{itemize}

We find that the distilled ensembles, which have better calibration, have greater AUC than individual models, and generally dominate them around the threshold corresponding to F1.

\section{Label smoothing in NER}\label{sec:label-smoothing}

We are not aware of a thorough study of the effects of label smoothing on NER tasks. Our experiments found that, similar to the NMT case, label smoothing did somewhat improve calibration for individual models. However, label smoothing gave mixed results when used in conjunction with our framework for ensembles and ensemble distillation, and generally the best results were achieved without it. We report our findings in \autoref{table:ner-labelsmooth}.

\begin{table*}[htb]
\centering
\scalebox{0.87}{
\begin{tabular}{c | c c c c c c }\toprule
Setting & \# Models & BS+ & BS- & B-BS & B-ECE & F1 \\
\midrule
Individual & Avg & 17.121 & 0.276 & 8.633 & 5.314 & 80.50 \\
LS Model & $\pm$ & 0.308 & 0.011 & 0.173 & 0.063 & 0.30 \\
\hline
\multirow{2}{*}{LS Distilled} 
& 3 & 14.312 & 0.217 & 8.301 & 5.101      & 81.96 \\
\multirow{2}{*}{non-LS Ensemble} 
& 6 & 14.355 & 0.224 & 8.557 & 5.915      & 82.09 \\
& 9 & 14.516 & 0.181 & 8.383 & 5.132      & 82.95 \\
\hline
\multirow{3}{*}{LS Ensemble} 
& 3 & 15.857 & 0.175 & 7.946 & 5.107      & 83.28 \\
& 6 & 15.366 & 0.179 & 7.915 & 5.150      & 83.21 \\
& 9 & 15.153 & 0.185 & 7.932 & 5.054      & 83.24 \\
\hline
\multirow{2}{*}{LS Distilled} 
& 3 & 15.012 & 0.214 & 8.415 & 5.324      & 82.04 \\
\multirow{2}{*}{LS Ensemble}
& 6 & 15.947 & 0.198 & 8.056 & 4.925      & 82.61 \\
& 9 & 15.876 & 0.231 & 8.445 & 4.955      & 81.50 \\
\bottomrule
\end{tabular}}
\caption{\label{table:ner-labelsmooth} Results for experiments on the CoNLL-2003 German dataset in which label smoothing was used. All models are have the IID architecture. Where applicable, the label smoothing factor $\alpha=0.1$.}
\end{table*}

\section{Dev results for CoNLL-2003}\label{sec:ner-dev}

\autoref{table:ner-dev} contains results for our ensemble and distilled ensemble experiments on the CoNLL-2003 English and German development splits. Each model is the same as the one used to produce the corresponding test result in \autoref{table:ner-results}.

\begin{table*}[h!]
\centering
\scalebox{0.87}{
\begin{tabular}{c c | c c c c c || c c c c c }\toprule
\multicolumn{2}{c}{}&\multicolumn{10}{c}{\textit{IID Model}} \\
\multicolumn{2}{c}{}& \multicolumn{5}{c}{\textit{English}}&\multicolumn{5}{c}{\textit{German}}\\
\midrule
Model & Setting & BS+ & BS- & B-BS & B-ECE & F1  & BS+ & BS- & B-BS & B-ECE & F1 \\ 
\midrule
\multirow{3}{*}{Ensemble}&3 & 3.154 & 0.095 & 2.546 & 0.905 & 95.30 & 12.235 & 0.183 & 6.950 & 4.514 & 86.39 \\
&6 & 2.878 & 0.087 & 2.333 & 0.850 & 95.79 & 11.694 & 0.185 & 6.888 & 4.420 & 86.75 \\
&9 & 2.801 & 0.089 & 2.303 & 0.745 & 95.72 & 11.688 & 0.177 & 7.006 & 4.279 & 86.76 \\ \midrule
\multirow{2}{*}{Ensemble}&3 & 3.377 & 0.119 & 2.884 & 1.706 & 95.02 & 11.890 & 0.235 & 7.610 & 5.872 & 85.37\\
\multirow{2}{*}{Distillation}&6 & 3.401 & 0.112 & 2.836 & 1.704 & 94.94 & 12.147 & 0.265 & 8.308 & 7.047 & 84.25 \\
&9 & 3.479 & 0.114 & 2.861 & 1.504 & 95.10 & 11.802 & 0.193 & 7.346 & 4.962 & 85.91 \\
\midrule
\multicolumn{2}{c}{}&\multicolumn{10}{c}{\textit{CRF Model}} \\
\multicolumn{2}{c}{}& \multicolumn{5}{c}{\textit{English}}&\multicolumn{5}{c}{\textit{German}}\\
\midrule
Model & Setting & BS+ & BS- & B-BS & B-ECE & F1  & BS+ & BS- & B-BS & B-ECE & F1 \\ 
\midrule
\multirow{3}{*}{Ensemble} &3 & 3.385 & 0.108 & 2.777 & 1.207 & 95.09 & 12.245 & 0.212 & 7.077 & 3.705 & 85.36 \\
&6 & 3.365 & 0.099 & 2.701 & 0.803 & 95.10 & 11.951 & 0.200 & 7.043 & 3.411 & 85.71 \\
&9 & 12.135 & 0.193 & 6.987 & 3.273 & 85.77 & 12.135 & 0.193 & 6.987 & 3.273 & 85.77 \\
\midrule
\multirow{2}{*}{Ensemble}&3 & 3.370 & 0.103 & 2.676 & 1.301 & 95.03 & 12.259 & 0.165 & 6.552 & 3.279 & 86.44 \\
\multirow{2}{*}{Distillation}&6 & 3.468 & 0.117 & 2.901 & 1.713 & 94.92 & 11.893 & 0.180 & 6.780 & 3.211 & 86.29 \\
&9 & 3.264 & 0.117 & 2.842 & 1.771 & 95.17 & 12.477 & 0.162 & 6.812 & 3.766 & 85.68 \\
\bottomrule
\end{tabular}
}
\caption{\label{table:ner-dev} Dev set results for the models reported in Table~\ref{table:ner-results}.}
\end{table*}

\section{Additional NMT results}\label{sec:ext-nmt}

\autoref{table:ext-nmt-results} gives results for ensembles of models which do not use SWA to combine checkpoints. We see that although the performance of the independent models is worse than those which use SWA, ensembles of them essentially match the performance of the corresponding ensembles which did use SWA. This suggests that ensembling obviates the need for checkpoint averaging. 

\begin{table}[ht]
\centering
\addtolength{\tabcolsep}{-1.7pt}
\begin{tabular}{@{}lcccc@{}}\toprule
{\bf Method}          & {\bf \# Models} & {\bf BLEU} & {\bf ECE-1} & {\bf ECE-5} \\ \midrule
\multirow{5}{*}{CE}       & 1               & 26.80        & 3.667       & 1.144      \\
                          & $\pm$           & 0.10         & 0.140       & 0.030      \\
                          & 3               & 28.38        & 0.904       & 0.303      \\
                          & 5               & 28.60        & 1.068       & 0.311      \\
                          & 7               & 28.42        & 1.286       & 0.325      \\
\bottomrule
\end{tabular}
\caption{\label{table:ext-nmt-results} Additional results for the WMT14 English $\rightarrow$ German task.}
\end{table}

\section{Information about datasets}\label{sec:datasets}

{\bf CoNLL-2003} comprises annotated text in two languages---English and German---taken from news articles. Details about sources, splits, and entity-type statistics can be found in~\cite{tjong-kim-sang-de-meulder-2003-introduction}. The NER information is annotated in IOB format; we modify this to IOB2 as a pre-processing step.

\vspace{5pt}
\noindent {\bf WMT16} gives parallel translations of parliamentary proceedings and news articles in a number of languages. We restrict our focus to the English-German language pair. Details about the corpus and splits can be found at~\url{http://www.statmt.org/wmt16/translation-task.html}. We follow the procedure provided by~\cite{ott-etal-2019-fairseq} for obtaining and processing the data.

\section{Information about computing infrastructure}\label{sec:computing}

{\bf NER.} The IID and CRF models for our NER experiments were each trained on one Nvidia GTX 1080 Ti GPU. Most of the models are trained using early stopping, which makes the training time somewhat variable, but typically requires 3-4 hours. Distilled student models tend to converge more quickly, sometimes requiring 2 hours or less to train.

\vspace{5pt}
\noindent {\bf NMT.} Training the Transformer models used for NMT is more computationally expensive due to the size of the training datasets. However, using 4 Nvidia RTX 2080 Ti GPUs, all models converged in less than 4 days, where we used 2 steps of gradient accumulation~\cite{ott-etal-2019-fairseq}. We note that it would be possible to reproduce our experiments using a single GPU by using more steps gradient accumulation, at the expense of longer training times.

\end{document}